\documentclass[12pt, a4paper]{article}

\usepackage[verbose=true,textheight=9in,
    textwidth=6.5in,
    top=1in,
    headheight=12pt,
    headsep=25pt,
    footskip=30pt]{geometry}

\usepackage{tikz, pgfplots}\usetikzlibrary{fadings}
\usetikzlibrary{shapes.multipart,  arrows.meta}
\usepackage{subcaption} \pgfplotsset{compat=newest}

\usepackage[utf8]{inputenc} 
\usepackage[T1]{fontenc}    
\usepackage{url}            
\usepackage{booktabs}       
\usepackage{amsfonts}       
\usepackage{nicefrac}       
\usepackage{xcolor}         
\usepackage{graphicx}
\usepackage{caption}
\usepackage[ruled,vlined]{algorithm2e}
\usepackage{amsmath}
\usepackage{amsthm} 

\newtheorem{definition}{Definition}[section]

\newtheorem{lemma}{Lemma}[section]

\newcommand\myeq{\mathrel{\stackrel{\makebox[0pt]{\mbox{\normalfont\tiny def}}}{=}}}

\title{Learning by the F-adjoint}

\author{%
 \sc  Ahmed Boughammoura\thanks{Correspondence to   ahmed.boughammoura@gmail.com}
   \\
{\small \sc Higher Institute of Informatics and Mathematics of Monastir, }\\ {\small\sc University of Monastir, 5000 Monastir, Tunisia.}  
  }

\begin{document}

\maketitle

\begin{abstract}
A recent paper by Boughammoura (2023) describes the back-propagation algorithm in terms of an alternative formulation called the F-adjoint method. In particular, by the F-adjoint algorithm the computation of the loss gradient, with respect to each weight within the network, is  straightforward and can simply be done.  In this work, we develop and investigate this theoretical framework to improve some supervised learning algorithm for feed-forward neural network. Our main result is that by
introducing some neural dynamical model combined by  the  gradient descent algorithm, we derived an equilibrium F-adjoint process which yields to some  local learning rule for deep feed-forward networks setting. Experimental results on  MNIST and Fashion-MNIST  datasets, demonstrate that the proposed approach provide a significant improvements on the standard back-propagation training  procedure \footnote{The code to reproduce the experimental results is
available at https://github.com/ahmadbougham/F-adjoint-Learning}. 
\end{abstract}

\section{Introduction}
~~~ 
It is well known that Artificial neural networks (ANNs) are inspired by biological  systems in brains and have been utilized in many applications, among others   classification, pattern recognition, and multivariate data analysis (Basheer and Hajmeer 2000 \cite{basheer2000artificial}). In practice, Back-propagation algorithm is the main component for gradient type learning rules for  (ANNs). In 1986,  D. Rumelhart and J. McCelland \cite{rumelhart1986learning} have presented the neural networks using  the back-propagation algorithm  to achieve some classification problems.  Since this work   many other were developed  to improve   both  theory and  practice  in association with the concept of back-propagation, see for example \cite{{whittington2019theories},{lillicrap2020backpropagation}} and references therein. 

Nevertheless, back-propagation which is generally used  in (ANNs) to provide the modification/update of their synaptic connections, or weights, during learning phase is revealed to be biologically unrealistic/non plausible. In particular, in (ANNs)  each synaptic weight update depends on the activity and computations of all downstream neurons, but biological neurons change their connection strength based solely on local signals  (see \cite{Mazzoni}). This could be mathematically expressed by the deficiency of local  formulation in the back-propagation method, as mentioned in \cite{Mazzoni}.  The main question that motivate this paper  is how one could reveal, capture and  specify a local behavior behind the  back-propagation procedure.

In the past few years, several models \cite{{bai2019deep},{millidge2020activation}, {whittington2019theories},{xie2003equivalence}} have been proposed that characterize  locally  the back-propagation. These models are considered to be biologically  plausible and  more closely approximate, in many applications, the back-propagation  algorithm. In this spirit of investigation, we have introduced the notion of F-adjoint propagation \cite{{boughammoura2023two},{boughammoura2023backpropagation}} to provide an alternative formulation of the back-propagation in order to highlight some mathematical properties  within this method. 

In the present work, we shall continue this exploration further by proposing a family of learning rules based on this F-adjoint method. In particular, by using a neural dynamical model, we provide an "asymptotic" version of the F-adjoint by which we  introduce some local learning rule to update the synaptic weights of each layer in an (ANN).

 Inspired by a recent gradient-based learning method for feed-forward neural networks
\cite{millidge2020activation}, we present learning rules that are (i) gradient-descent based algorithm, (ii) local in space, (iii) local in time, and in particular do not require storing intermediate states. To meet all these criteria at once, we start from gradient-descent optimization of the loss function, and we reformulate   learning process within the F-adjoint formulation. In particular, we provide the equivalence of the backpropagation and some  methods based on the so-called equilibrium propagation \cite{millidge2020activation}  which is essentially a combination of a  neural dynamics with a learning rule based on synaptic plasticity models, which provides the change in the synaptic weights.

To end this section, let us emphasize that the primary  goal of this work is the development of a significant mathematical  framework in which we introduce some precise definitions that cover the concepts commonly needed in supervised learning algorithm for feedforward neural network, such as sequential models, forward/F-propagation  and backward/F-adjoint propagation and local learning rule, etc. Specifically, we could be able, within  this F-adjoint formulation, to provide and prove simply, many  properties about the backpropagation method  in both fundamental and application viewpoints.

\section{Notation and mathematical background} \label{sec:bg}
~~~ 
The purpose of this section is to provide a brief overview of the   mathematical framework  used in the F-adjoint theory which is introduced in \cite{boughammoura2023backpropagation}
We consider the simple case of a fully-connected deep multi-layer perceptron (MLP) composed of $L$ layers trained in a supervised setting.  Following \cite{baldi2021} (see (2.18) in page 24), we will denote such an architecture by
\begin{equation}
\label{architectue}
A[N_0, \cdots, N_\ell,\cdots, N_L]
\end{equation}
where $N_0$ is the size of the input layer, $N_\ell$ is the size of hidden layer $\ell$,
and $N_L$ is the size of the output layer; $L$ is defined as the depth of the ANN, then the neural network is called as Deep Neural Network (DNN).
 
\begin{table}[htp]
\centering
\def\arraystretch{1.2}
 
\small 
        \begin{tabular}{|l|l|}
 \hline
    Term & Description \\
 \hline
   
    $W^{\ell}$ & Weight matrix of the layer ${\ell}$ with bias,  $W^{\ell}\in\mathbb{R}^{N_{\ell}\times (N_{\ell-1}+1)}$\\[5pt]
     \hline
 $ W_\sharp^{\ell}$ & Weight matrix of the layer ${\ell}$ without bias,  $W^{\ell}\in\mathbb{R}^{N_{\ell}\times N_{\ell-1}}$\\[5pt] 
     \hline
    $Y^{\ell}$ & Pre-activation vector at layer ${\ell}$, $Y^{\ell} = W^{\ell}X^{\ell-1}\in\mathbb{R}^{N_{\ell}}$\\[5pt]
 \hline
    $X^{\ell}$ & Activation vector at the layer ${\ell}$, $X^{\ell} =\left(\sigma^{\ell}(Y^{\ell}),1\right)^\top\in\mathbb{R}^{N_{\ell}}\times\{ 1\}$\\[5pt] 
     \hline
    $\sigma^\ell$ & Point-wise activation function of the layer ${\ell}$ with bias,   $\sigma^\ell :\mathbb{R}^{N_{\ell}}\ni Y^{\ell}\longmapsto\sigma^{\ell}(Y^{\ell})\in\mathbb{R}^{N_{\ell}}$\\[5pt]
   \hline
\end{tabular}

\caption{Notations describing a (DNN).}\label{notation}
\end{table}  


\medskip
~~~
Notation describing a multi-layered FF network (MLP) is summarized in table \ref{notation}. To simplify notation, we assume bias is stored in an extra column on each weight matrix $W^{\ell}$ and we assume a $1$ is concatenated to the end of the   activity vector $X^{\ell}$. Without loss of generality, we may assume that $\sigma^\ell=\sigma$ for all ${\ell}\in\{1,\cdots, L\}$.
Consider an $L$-layer feedforward network with input $X^0\in\mathbb{R}^{N_0}$.
 The standard training approach is to minimize a loss, $J(y, \hat{y} )$, with respect to the
weights, where $y$ is the desired output and  $\hat{y}=X^L$ is the prediction of the network.
\section{The F-propagation and F-adjoint}\label{B}

~~~ Modern feedforward deep networks are built on the key concept of layers. In the forward passes, each network consists of a family of some $L$ pre-activation and activation vectors, where $L$ is the depth of the network. To update these networks, the backward passes rely on backpropagating through the same $L$ layers via some rule, which generally necessitates that we store the intermediate values of these layers. Thus,   sequence mathematical models are commonly used  for many applications of deep networks. Moreover, sequence modeling has shown continuous advances in model architectures.  Specifically, a   feedforward sequence model can be written as the following. 
\begin{definition}[Sequence model]\label{seqdef1}
{}\noindent

 Suppose that we are given an input vector $X^0=(X^0_1,\ldots,X^0_m)$,
and try to predict some  output vector $y^0=(y^0_1,\ldots,y^0_m)$. To predict the output $y^0_i$ for some  $i$, we are
constrained to only use  $X^0_1,\ldots,X^0_i$.  Mathematically, a sequence model  is a function ${f :
\mathcal{X}^{m} \rightarrow \mathcal{Y}^{m}}$ that produces the mapping
\begin{equation}
\hat{y}^0 = f(X^0)
\end{equation}
if it satisfies the causal constraint that $\hat{y}^0_i$ depends only on
${X^0_1,\ldots,X^0_i}$ and not on any  ${X^0_{i+1},\ldots,X^0_m}$. In addition, 
when $X^0_i$, for some $i$, is computed via two-step, we shall refer to this model as  a two-step sequence model.
\end{definition}

Let us noted that, the main goal of learning in the sequence model  setting is to find a function $f$ that minimizes some expected loss between the actual outputs and the predictions, $L(y^0, f(X^0))$. Within this  formalism and based on the   idea of the two-scale rule for back-propagation  \cite{boughammoura2023two}, the first author  have introduced the F-adjoint concept in 
\cite{boughammoura2023backpropagation}.

\begin{figure}[htp]
\centering
\begin{tikzpicture}[A/.style={scale=0.8, transform shape},
    B/.style={scale=0.7, transform shape},
    every edge/.style={->, draw}]
    \node[A] (A1) at (0, 2.3) {$Y^\ell =W^\ell X^{\ell-1}$};
    \node[A] (A2) at (0, 1) {$X^{\ell-1}_*=(W_\sharp^\ell)^\top Y^\ell_*$};
    \node[A] (A4) at (5, 1) {$Y^\ell_*=X^{\ell}_*\odot \sigma'(Y^\ell)$};
    \node[A] (A5) at (5, 2.3) {$X^\ell=\sigma(Y^\ell)$};
    \draw (A1) edge node[B, left] {$\mathrm{F}$-adjoint} (A2);
    \draw (A4.west) edge[bend left=10] node[ below] {$\ \ {\mathrm{F}_*}$-pass} (A2.east);
    \draw (A5) edge node[B, right] {$\mathrm{F}$-adjoint} (A4);
    \draw (A1.east) edge[bend left=10] node[above]{{${\mathrm{F}}$-pass}} (A5.west);
\end{tikzpicture}

\vspace*{2pt}
\caption{A schematic diagram showing the $\mathrm{F}$ and $\mathrm{F}_*$ processes.}
\label{fig:Fpasses}\end{figure}
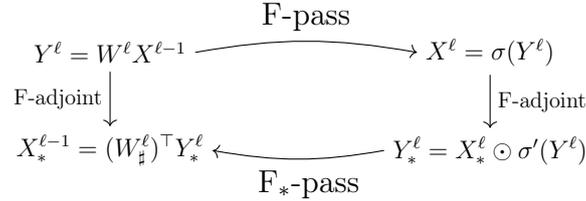

 For the sake of coherency  of presentation we shall recall and revise the  definition of the this notion and provide some straightforward properties and improvements of this adjoint-like representation.
\begin{definition}[An F-propagation]\label{def1}
{}\noindent

Le $X^0\in\mathbb{R}^{N_0}$ be a given data, $\sigma$ be a coordinate-wise map from $\mathbb{R}^{N_\ell}$ into $\mathbb{R}^{N_{\ell+1}}$ and $W^{\ell}\in \mathbb{R}^{{N_{\ell}}\times{N_{\ell-1}}}$ for all ${1\leq \ell\leq L}$. We say that we have a two-step recursive F-propagation   $F$  through the DNN $A[N_0,\cdots, N_L]$ if   one has the following family of vectors
\begin{equation}
\label{F-def}
F(X^0):=\begin{Bmatrix}
Y^{1},X^{1},\cdots,X^{L-1},  Y^{L},  X^{L}
\end{Bmatrix}
\end{equation} with
\begin{equation}
\label{F-def-eq1}
Y^\ell=W^{\ell}X^{\ell-1}, \ X^\ell=\sigma(Y^\ell),\ X^\ell\in\mathbb{R}^{N_\ell},\ \ell=1,\cdots, L.
\end{equation}
\end{definition}
Before going further, let us point that in the above definition the prefix "F" stands for "Feed-forward".
\begin{definition}[The F-adjoint of an F-propagation]\label{def2}
{}\noindent

Let $(X^0,y)\in\mathbb{R}^{N_0}\times \mathbb{R}^{N_L}$ be a given feature-target pair and let $X^{L}_{*}\in\mathbb{R}^{N_L}$ be a given vector.  We define the F-adjoint     ${F}_{*}$, through the DNN $A[N_0,\cdots, N_L]$, associated to the F-propagation  $F(X^0)$  as follows
\begin{equation}
\label{F*-def}
F_{*}(X^{0},y):=\begin{Bmatrix}X^{L}_{*},
Y^{L}_{*}, X^{L-1}_{*},\cdots, X^{1}_{*},Y^{1}_{*}
\end{Bmatrix}
\end{equation} with
\begin{equation}
\label{F*-def-eq1}
Y^\ell_{*}=X^{\ell}_{*}\odot {\sigma}'(Y^\ell), \ X^{\ell-1}_{*}=(W_\sharp^\ell)^\top Y^\ell_{*},\ h=L,\cdots, 1.
\end{equation}
\end{definition}
Firstly, let us precise that we  shall introduce the following schematic diagram to illustrate how to to recover/find easily the formulas \eqref{F*-def-eq1} (see Figure\ref{fig:Fpasses}). 

To refine and develop further the $F$adjoint transformation we shall introduce the  $F$-adjoint through a single $\ell^{\mathrm{th}}$ hidden layer  

Secondly, we shall mention that one gets the following immediate properties gives some straightforward relations  between the  vectors
$Y^\ell_*$,    $X^{\ell-1}_*$, $Y^\ell$ and    $X^{\ell-1}$.

Hereafter we us the following notations  
 $$F(X^0)\myeq\left\{\left(Y^\ell, X^\ell\right) \right\}_{\ell=1}^L\ \mathrm{et} \ F_*(X^0,y)\myeq\left\{\left(X_*^{\ell}, Y_*^{\ell}\right) \right\}_{\ell=L}^1$$ 
 the set of steps taken during forward propagation and back-propagation associated with the input $X^0$ through the DNN $A[N_0,\cdots, N_L]$.

\subsection{Properties of the F-adjoint}
\begin{lemma}
\noindent{}\\
 For a  fixed data point  $(x, y) \in \mathbb{R}^{N_0}\times\mathbb{R}^{N_L}$, with feature vector $x$ and label $y$ and  a  fixed loss function $J$.  If $X^{L}_{*}=\frac{\partial J}{\partial X^{L}}$ then for any $\ell\in\{1,\cdots, L\}$, we have    
 \begin{equation}\label{eq00:properties}
 {Y_{*}^{\ell}}\left({X^{\ell -1}}\right)^\top=  \frac{\partial J}{\partial W^{\ell}}.
 \end{equation} 
\end{lemma}
The formulas given by \eqref{eq00:properties} have two factors:
the F-adjoint vector $ Y^{\ell}_{*} $ through the  layer ${\ell}$ and  the transpose of the activation vector $\left(X^{\ell-1}\right)^\top$ through   the   precedent   layer. Both terms are available in $F_{*}(X^{0},y)$ and $F(X^{0})$ respectively. Thus, to update the weight  $W^{\ell}$ of layer $\ell$ we need   the couple of vectors
$\left( Y^{\ell}_{*},X^{\ell-1}\right)\in F_{*}(X^{0},y)\times F(X^{0})$ for any $\ell\in\{1,\cdots, L\}$. As consequence, to evaluate the loss gradient we  simply  need to determine only  the sets  $ F(X^{0})$ and $F_{*}(X^{0},y)$.
 
 \begin{proof}
\noindent{}\\
Firstly, let us recall that if one has
\begin{equation}\label{eq00} W\in \mathbb{R}^{n\times q},\ X\in \mathbb{R}^{(q-1)}\times\{1\} ,\  Y=WX, \ f=f(Y)
\end{equation}
then\begin{equation}\label{eq000} \frac{\partial f}{\partial W}\myeq\left(\frac{\partial f}{\partial W_{i,j}}\right)_{i,j} =\frac{\partial f}{\partial Y}X^\top\in \mathbb{R}^{n\times q} 
\end{equation} and
 \begin{equation}\label{eq000}  \frac{\partial f}{\partial X}\myeq \left(\frac{\partial f}{\partial X_i}\right)_i=W_\sharp^\top\frac{\partial f}{\partial Y}\in \mathbb{R}^{q\times 1}
\end{equation} since $X_q=1$.
Furthermore, for any   differentiable function  with respect to  $Y$
\begin{equation*}
F:\ \mathbb{R}^n\ni Y\mapsto X:=\sigma(Y)\in \mathbb{R}^{n}\mapsto F(X) \in \mathbb{R}
\end{equation*} we have
	\begin{equation}\label{chain:2}
	 \frac{\partial F}{\partial Y} =
				 \frac{\partial F}{\partial X}\odot\sigma'(Y).
	\end{equation}	

Now, by using the chain rule for the loss function $J$ with respect to $ Y^\ell$, $ X^{\ell-1}$  and $ W^\ell$ respectively, we have 
$$\frac{\partial J}{\partial Y^\ell}=\frac{\partial J}{\partial X^\ell}
\frac{\partial X^\ell}{\partial Y^\ell}= X^{\ell}_*\odot\sigma'(Y^\ell)=Y_*^\ell, $$
and
$$ 
\frac{\partial J}{\partial X^{\ell-1}}=\left(W_\sharp^\ell\right)^\top \frac{\partial J}{\partial Y^\ell}=\left(W_\sharp^\ell\right)^\top Y^\ell_*=X^{\ell-1}_*
$$
As consequence 
$$
\frac{\partial J}{\partial W^\ell}=\frac{\partial J}{\partial {Y^\ell}} \left(X^{\ell-1}\right)^\top
=Y^\ell_*\left(X^{\ell-1}\right)^\top.$$
\end{proof}

\subsection{Learning by F-adjoint}
~~~
The mathematical learning rule to change a weight of a network can be local or non-local. According to a 'neuro-physiological postulate' which is referred to as Hebb’s rule, the second possibility must be excluded in case the weight is associated with a synapse. More precisely, 
 in \cite{Hebb},   Hebb conjectured that "When an axon of cell $A$ is near enough to excite a cell $B$ and repeatedly or persistently takes part in firing it, some growth process or metabolic change takes place in one or both cells so that $A$'s efficiency, as one of the cells firing $B$, is increased". 
Thus, a change of the weights may depend only on  local, in space and time, variables.  In this context,  we shall introduce the following mathematical definition of a  local  learning rule written with the F-adjoint formulation.
 
\begin{definition}[Local F-learning rule]\label{def1}
{}\noindent

For a given  F-F${}_*$ sets
$$F(X^0)\myeq\left\{\left(Y^\ell, X^\ell\right) \right\}_{\ell=1}^L\ \mathrm{et} \ F_*(X^0,y)\myeq\left\{\left(X_*^{\ell}, Y_*^{\ell}\right) \right\}_{\ell=L}^1$$ 
We shall say that the associated deep feedforward network is trained by a local F-learning rule if every $Y_*^{\ell},\ \ell = L,\cdots , 1$ depend only on the initial data $X_*^{L}$ and the F-propagation set $F(X^0)$.
\end{definition}

\subsection{Non-local learning  rule}
~~~
Consider a dataset $\mathcal{D} = \{\mathbf{X},\mathbf{Y}\}$,  for any    data point  $(x, y) \in \mathbb{R}^{N_0}\times\mathbb{R}^{N_L}$, with feature vector $x$ and label $y$ and  a  fixed loss function $J$.  If $X^{L}_{*}=\frac{\partial J}{\partial X^{L}}$ then for any $\ell\in\{1,\cdots, L\}$, we have   the following learning  update rule given by \eqref{eq00:properties}, namely
$$
 {Y_{*}^{\ell}}\left({X^{\ell -1}}\right)^\top=  \frac{\partial J}{\partial W^{\ell}}.
$$
 At iteration $t_0$, an explicit SGD iteration subtracts the gradient of the loss function $J$ from the parameters: $$W^\ell{(t_0+1)} = W^\ell{(t_0)} - \eta \frac{\partial J}{\partial W^\ell{(t_0)}}= W^\ell{(t_0)} - \eta {Y_{*}^{\ell}(t_0)}\left({X^{\ell -1}(t_0)}\right)^\top,$$ with step size $\eta$. The back-propagation \cite{rumelhart1986learning} is a common algorithm for implementing SGD in deep networks. Here, a pseudo-code of the SGD rewritten in the F-adjoint formulation.

\begin{algorithm}[H]
\SetAlgoLined
\KwData{Dataset $\mathcal{D} = \{\mathbf{X},\mathbf{Y}\}$, parameters $W = \{W^1, \cdots, W^L\}$,  learning rate $\eta$.} 
\tcc{Iterate over dataset}
\For{$(x, y \in \mathcal{D})$}{
\tcc{F-propagation}
$X^0=x$\\ 
\For{$\ell=1,\cdots, L$}{
$Y^\ell=W^\ell X^{\ell-1}$\\
$X^{\ell} = \sigma(Y^\ell)$ 
}
\tcc{F-adjoint propagation}
$X_*^0=\frac{\partial J}{\partial X^L}(., y)$\\ 
\For{$\ell=L,\cdots, 1$}{
$Y_*^\ell=X_*^{\ell}\odot\sigma'(Y^\ell)$\\
$X_*^{\ell-1} =(W_\sharp^\ell)^\top Y_*^\ell$ 
}
\tcc{Update weights}
\For{$\ell=L,\cdots, 1$}{
${W^\ell} = {W^\ell} - \eta {Y_*^\ell}(X^{\ell -1})^\top $
}
}
\caption{Nonlocal F-adjoint algorithm}\label{algo-nonlocal}
\end{algorithm}

\subsection{Local learning  rule}
~~~
We now present an explanation of  the main idea behind  the  local learning rule by using the F-adjoint terms $Y_*^{\ell}, X_*^{\ell}$.  As it is clear from the above non-local learning rule (see Algorithm 1),  the major disadvantage inherent in the use of  this method lies in the  difficulty of computing the factor 
\begin{equation}\label{ll+0}
Y_*^{\ell}\myeq X^{\ell}_*\odot\sigma'(Y^\ell).
\end{equation} 
In fact, the computation of this term  is achieved by  applying, recursively from the output loss through the layer ${l+1}$,  the following relation  
\begin{equation}\label{ll+1} X^{\ell}_*= \left(W_\sharp^{\ell +1}\right)^\top Y^{\ell+1}_*= \left(W_\sharp^{\ell +1}\right)^\top X^{\ell +1}_*\odot\sigma'(Y^{\ell+1}).
\end{equation}
Since the update at each $\ell$-layer depends on the F-adjoint terms of all superordinate layers in the hierarchy, thus  this procedure is not local. 

\textbf{Main idea of this approach}
The procedure does not compute the true gradient of the objective function, but rather
approximates it at a precision which is proven to be directly
related to the degree of symmetry of the feedforward and
feedback weights.

Here, we shall propose a method for computing the F-adjoint term $ X^{\ell}_*$
using a dynamical systems approach. Recall that
  $X^{\ell}_*=\frac{\partial J}{\partial X^l}$ (The loss derivative with respect to the $\ell$-layer activation). Particularly, we define a dynamical system on  a   space state similar to the F-set $F(X^0)$ denoted 
  $\left\{\left(X_\sim^\ell(t), Y_\sim^\ell(t)\right)\right\}_{\ell=L}^1$, where of each $\ell$-layer the  pair $\left(X_\sim^\ell(t), Y_\sim^\ell(t)\right)$ corresponds to  the  state, at time $t$,  of the dynamical system  driven by  the input $ X^{\ell}_*(t_0)$ ( $t_0$ is a   fixed iteration step) and  starting from $ X^{\ell}(t_0)$.  The simplest dynamical system to achieve this is a leaky-integrator driven by top-down feedback, defined on the  state space of  F-propagation type,  by the following autonomous    leaky integrate-and-fire neuron form, introduced in \cite{gerstner2002spiking} ( Chapter 4.).
\begin{equation}\label{sys1}
  \left\{\begin{array}{ll} 
   \frac{d {X_\sim^{\ell}}}{dt}(t) &= -{X_\sim^{\ell}}(t) +  X^{\ell}_*(t_0) \\[5pt]
   {X_\sim^{\ell}}(0) &=  X^{\ell}(t_0)
\end{array}\right. 
\end{equation} where the F-F${}_*$ pair $\left(X^{\ell}(t_0),X^{\ell}_*(t_0)\right) $ is given at a fixed  iteration  $t_0>0$ for the $\ell$-hidden layer. 
the above dynamical system that is defined by an F-pass vector ${X_\sim^{\ell}}$ will be  called an F-dynamical system.

 Then \eqref{sys1} converges to the minimum of some state ${X_\sim^{\ell}}(\infty)$ which is given by $ \frac{d {X_\sim^{\ell}}}{dt}(t)=0$,  thus
\begin{equation}\label{sys11}
{X_\sim^{\ell}}(\infty)=X^{\ell}_*(t_0)
\end{equation}
The equality \eqref{sys11} shows clearly that the steady-state equilibrium of \eqref{sys1} is exactly equal to necessary  term needed to update  the weights.

 On the other hand, by \eqref{ll+1} and \eqref{sys11} one has
 \begin{equation}\label{sys12}
  \left\{\begin{array}{ll} 
   \frac{d {X_\sim^{\ell}}}{dt}(t) &= -{X_\sim^{\ell}}(t) + \left(W_\sharp^{\ell +1}(t_0)\right)^\top X^{\ell +1}_*(t_0)\odot\sigma'(Y^{\ell+1}(t_0)) \\[5pt]
 {}  & =  -{X_\sim^{\ell}}(t) + \left(W_\sharp^{\ell +1}(t_0)\right)^\top\left( {X_\sim^{\ell +1}}(\infty)\right)\odot\sigma'(Y^{\ell+1}(t_0)) \\[5pt]    
   {X_\sim^{\ell}}(0) &=  X^{\ell}(t_0)
\end{array}\right. 
\end{equation}
   Let us denote  for any $\ell=1,\cdots, L$
 \begin{equation}\label{sys122}
 {Y_\sim^{\ell}}(\infty)=
 \left( {X_\sim^{\ell}}(\infty)\right)\odot\sigma'(Y^{\ell}(t_0))
 \end{equation} and as consequence, we shall define the  equilibrium F-propagation $ {F_\sim}(X^0,y)$, associated to the pair of feature and target $(X^0,y)$ as follows :
$$ {F_\sim}(X^0,y)\myeq\left\{\left( {X_\sim^{\ell}}(\infty), {Y_\sim^{\ell}}(\infty)\right) \right\}_{\ell=L}^1$$

Then \eqref{sys12} is rewritten as
 \begin{equation}\label{sys122}
 \left\{\begin{array}{ll} \frac{d {X_\sim^{\ell}}}{dt}(t) &=  -{X_\sim^{\ell}}(t) + \left(W_\sharp^{\ell +1}(t_0)\right)^\top\left({Y_\sim^{\ell+1}}(\infty)\right)
 \\[7pt]
   X_\sim^{\ell}(0) &=  X^{\ell}(t_0)\end{array}\right. 
 \end{equation} 

%

Let us emphasize that  the systems \eqref{sys122}  and \eqref{sys1} are equivalent, thus they   converges to the same optimum, namely $X^{\ell}_*(t_0)$. 

The equivalent F-dynamical system \eqref{sys122} forms the backbone of the local F-adjoint learning rule. The algorithm \ref{algo-local} proceeds as follows. First, an F-pass computes the $L$ couples constituting the set $F(X^0)$ then starting from the top-layer equilibrium F-propagation pair $\left(X_\sim^{L}(\infty)=X^L_*,Y_\sim^{L}(\infty)=X_\sim^{L}(\infty)\odot \sigma'(Y^L)\right)$  the equilibrium state $X_\sim^{L-1}(\infty)$ of the penultimate layer is computed,  then the associated 
$Y_\sim^{L-1}(\infty)=X_\sim^{L-1}(\infty)\odot\sigma'(Y^{L-1})$, the network enters into an F-dynamical  phase where  the system \eqref{sys122} is iterated  for all layers until convergence for each layer. 
Upon convergence, the equilibrium state of each layer multiplied by $ \sigma'(Y^{\ell})$ is precisely equal to  $Y_*^{\ell}$, and are used to update the weights.

\vspace*{0.5cm}

\begin{algorithm}[H]
\SetAlgoLined
\KwData{Dataset $\mathcal{D} = \{\mathbf{X},\mathbf{Y}\}$, parameters $W = \{W^1, \cdots, W^L\}$, inference learning rate $\tau$, weight learning rate $\eta$.} 
\tcc{Iterate over dataset}
\For{$(x, y \in \mathcal{D})$}{
\tcc{F-propagation}
$X^0=x$\\ 
\For{$\ell=1,\cdots, L$}{
$Y^\ell=W^\ell X^{\ell-1}$\\
$X^{\ell} = \sigma(Y^\ell)$ 
}
\tcc{Equilibrium F-adjoint}
$X_\sim^{L}(\infty)=X^L_*:= \frac{\partial J}{\partial X^L}(., y)$\\
$Y_\sim^{L}(\infty)=X_\sim^{L}(\infty)\odot \sigma'(Y^L)$\\
\For{$\ell=L,\cdots, 1$}{\tcc{Equilibrium state $X_\sim^{\ell-1}(\infty)$} 
$ {X_\sim^{\ell-1}} =  X^{\ell-1}$\\
\While{not converged}{
$ dX_\sim^{\ell-1}(t)= - X_\sim^{\ell-1}(t) +\left(W_\sharp^{\ell}\right)^\top  Y_\sim^{\ell}(\infty)  $ \\
{$X_\sim^{\ell-1}(t+1) = X_\sim^{\ell-1}(t)+\tau  dX_\sim^{\ell-1}(t)$}\\
}
$Y_\sim^{\ell-1}(\infty)= \left(X_\sim^{\ell-1}(\infty)\right)\odot\sigma'(Y^{\ell-1})$}
\tcc{Update weights}
\For{$\ell=L,\cdots, 1$}{
${W^\ell} = {W^\ell}  - \eta Y_\sim^\ell(\infty)(X^{\ell -1})^\top $ 
}
}
\caption{Local F-adjoint   algorithm}\label{algo-local}
\end{algorithm}
%
%

\bigskip

Let us emphasize that the learning update rule in the above algorithm is given by
 $\eta Y_\sim^\ell(\infty)(X^{\ell -1})^\top$, thus the associated learning rule is

\begin{equation}\label{local-learning}
\frac{\partial J}{\partial W^{\ell}}= Y_\sim^\ell(\infty)(X^{\ell -1})^\top 
 \end{equation} 
where for each $\ell\in\{ L,\cdots, 1\}$ one has
$$
Y_\sim^{\ell}(\infty)= \left(X_\sim^{\ell}(\infty)\right)\odot\sigma'(Y^{\ell}),\  X_\sim^{L}(\infty)=X^L_*
$$ which confirm that the present learning rule depend only on the initial data $X_*^{L}$ and the F-propagation set $F(X^0)$, thus it is a local one   relatively to the Definition \ref{def1}.
\section{Experiments}
~~~ 
Firstly, let us recall that the  MNIST (Modiﬁed National Institute of Standards and Technology) is a standard benchmark dataset of handwritten digit images made in 1998 (Lecun et al., \cite{lecun1998gradient}). Firstly, we shall flattened the 28*28 images to have the following model architecture.

Secondly, we shall compare performance of the proposed learning algorithms
on two image classification datasets (MNIST  and Fashion-MNIST) by using  a simple shallow  Multilayer Perceptron (MLP) with the architecture   $A[784, 128, 10]$,  sigmoid activation function through each hidden layer, Xavier weight initialization, Cross-entropy loss function and learning rate $0.001$. Let us mention, that we have deliberately fixed this (MLP)  architecture for all  conducted experiments.

Thirdly, for the optimizer, here we choose to use the standard SGD algorithm written with the F-adjoint and equilibrium F-adjoint steps respectively. We train the model by batch with the following appropriate parameters :
number of epochs $= 1000$, learning rate $= 0.001$ and batch size $= 128$. A summary of parameter choices are given in the following table.

Finally, in order to determine the performance of the neural network and predictions and report the results produced by our neural network, we choose to calculate the training accuracy and testing accuracy.
\subsection{Results for MNIST dataset}

 \begin{table}[htb]
\centering
\def\arraystretch{1.2}
 
        \begin{tabular}{|l|c|c|}
 \hline  
F-adjoint model & Training accuracy          & Testing  accuracy     \\[5pt]
   \hline   
Non-local learning rule     &  $99.233\ \% $ &  $97.879\ \% $\\[5pt]
   \hline  
 Local learning rule     &  $97.094\ \% $ &  $96.629\ \% $\\[5pt]
   \hline     
\end{tabular}
\vspace*{5pt}
\caption{Accuracy results for the MNIST dataset. \label{tab:notation}}
\end{table}

\subsection{Results for Fashion-MNIST dataset}
 
 \begin{table}[htb]
\centering
\def\arraystretch{1.2}
 
        \begin{tabular}{|l|c|c|}
 \hline  
F-adjoint model & Training accuracy          & Testing  accuracy     \\[5pt]
   \hline   
Non-local learning rule     &  $95.025\ \% $ &  $89.770\ \% $\\[5pt]
   \hline
Local learning rule     &  $89.893\ \% $ &  $88.58 \ \% $\\[5pt]
   \hline           
\end{tabular}
\vspace*{5pt}
\caption{Accuracy results for the Fashion-MNIST dataset. \label{tab:notation}}
\end{table}

\begin{figure}[htb]
  \centering

  \begin{minipage}[b]{0.47\textwidth}
    \centering
    \includegraphics[width=\textwidth]{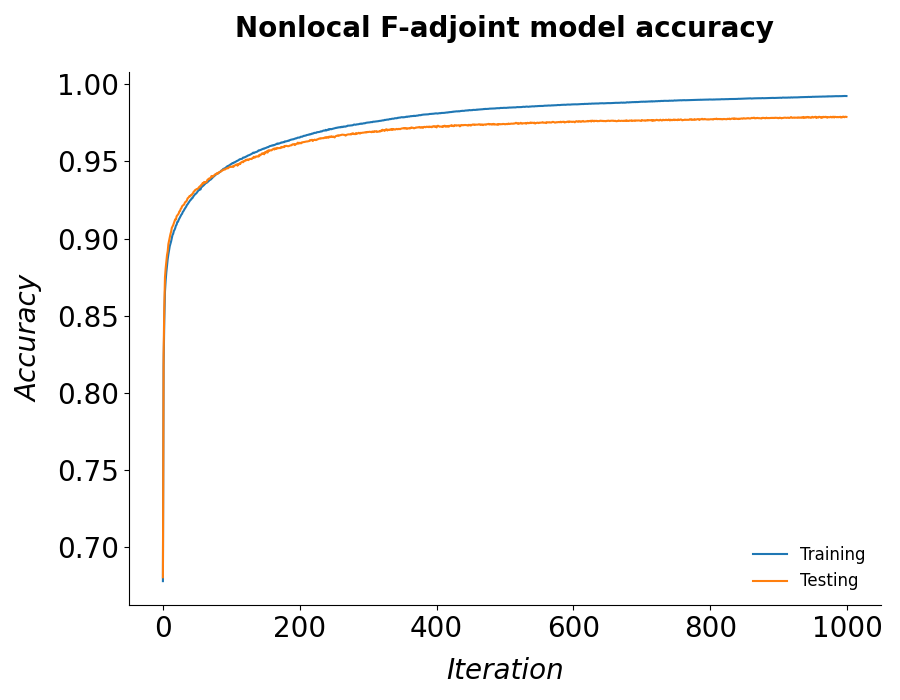}
    \caption{Accuracy for MNIST  with nonlocal learning rule. }
    \label{fig:figure1}
  \end{minipage}
  \hfill
  \begin{minipage}[b]{0.47\textwidth}
    \centering
    \includegraphics[width=\textwidth]{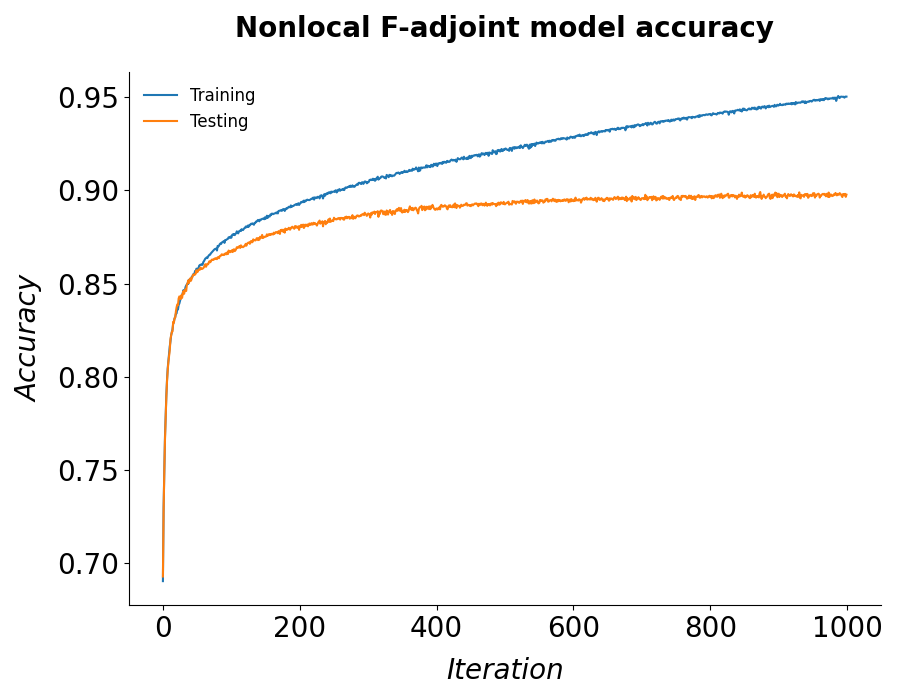}
    \caption{Accuracy for Fashion-MNIST   with  nonlocal learning rule. }
    \label{fig:figure2}
  \end{minipage}

\vspace*{.5cm}
  \begin{minipage}[b]{0.47\textwidth}
    \centering
    \includegraphics[width=\textwidth]{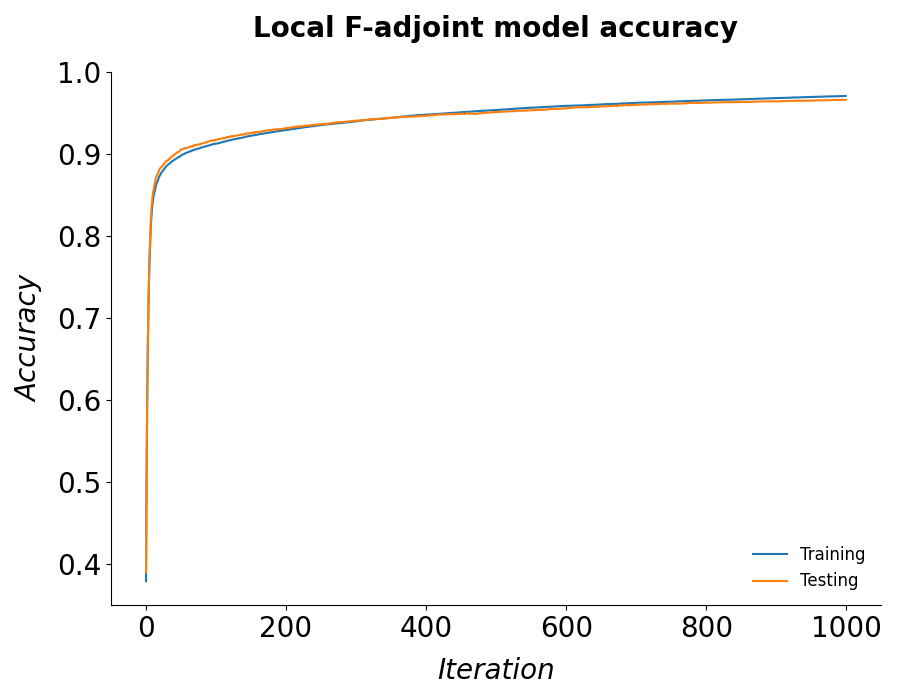}
    \caption{Accuracy for MNIST  with local learning rule.
}
    \label{fig:figure1}
  \end{minipage}
  \hfill
  \begin{minipage}[b]{0.47\textwidth}
    \centering
    \includegraphics[width=\textwidth]{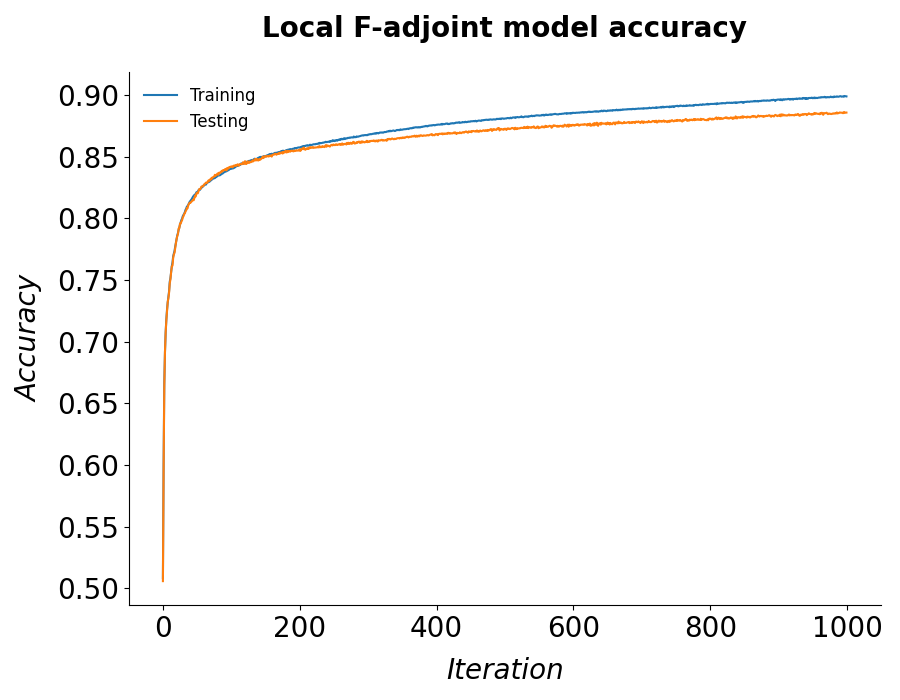}
    \caption{Accuracy for Fashion-MNIST   with local learning rule. }
    \label{fig:figure2}
  \end{minipage}
\vspace*{.5cm}
  
\end{figure}

It can be seen that  training   accuracy tends to increase as the  epoch increases.   Meanwhile, testing accuracy rate didn't increase significantly.

In this section, we test and compare F-adjoint based algorithms to  by training these models on supervised classification tasks. Results  

\section{Conclusion}\label{sec:lim}
~~~
In this work, by using the F-adjoint formulation combined with a leak-integrator dynamical system, we propose a local F-learning rule to   train feed-forward deep neural networks. The proposed model consists only of two steps, namely F-propagation  and an equilibrium   F-propagation  phase. Experiment results revealed that the proposed approach
can achieve   results similar to the one based only on the F-adjoint version, which is equivalent to the standard back-propagation method. We believe
that F-adjoint techniques provide some mathematical setting within we can describe straightforward some important training processes.

 Our work suggests an immediate direction of future research, including:

- Replace the $W^T$ weight transpose by a random matrix B with the same dimensions as $W^T$ for each layer in the F-dynamical system \eqref{sys122}. This procedure yields to  some new variation of the so-called feedback alignment algorithm \cite{Fa};

- For the equilibrium F-adjoint, we propose some scaling on the  initialization of the procedure by choosing $\mu X^L_\sim (\infty)$  for some $\mu>0$ instead of $X^L_\sim (\infty)$;

- Investigate the regularization process in the local learning rule.

  However, some open questions also remain on the side of mathematical generalization. F-adjoint framework only describes propagation processes in feed-forward artificial neural networks, and it may be different in different contexts. A particularly interesting type of result to consider in the future regards the relationship between F-adjoint concept and some recurrent-type artificial neural networks, for example. 
  
  To end this section, we may conjecture that  by \emph{local} learning rule \eqref{local-learning}, the classification accuracies tends to be  approximately similar for the training and the testing datasets.

\bibliographystyle{plain}

\end{document}